\documentclass[runningheads]{llncs}
\usepackage[T1]{fontenc}
\usepackage{hyperref}
\usepackage{tikz}
\usepackage{comment}
\usepackage{amsmath,amssymb}
\usepackage{color}

\usepackage{xspace}
\usepackage{multirow}
\usepackage{caption}
\usepackage{subcaption}
\usepackage{cite}

\newcommand{\ie}{\emph{i.e.,}\xspace}
\newcommand{\wrt}{\emph{w.r.t.}\xspace}

\newcommand{\eg}{\emph{e.g.,}\xspace}

\newcommand{\featvec}[3]{\ensuremath{#1_{[#2,#3]}}}
\newcommand{\partmap}[2]{\ensuremath{P(k^{(#1)},#2)}}
\newcommand{\isep}{\mathrel{{.}\,{.}}\nobreak}

\begin{document}
\title{PARTICUL: Part Identification with Confidence measure using Unsupervised Learning}
\titlerunning{Part Identification with Confidence measure using Unsupervised Learning}
\author{Romain Xu-Darme\inst{1,2} \and
Georges Quénot\inst{2} \and Zakaria Chihani\inst{1} \and
Marie-Christine Rousset\inst{2}}
\authorrunning{R. Xu-Darme et al.}
\institute{Université Paris-Saclay, CEA, List, F-91120, Palaiseau, France \and Univ. Grenoble Alpes, CNRS, Grenoble INP, LIG, F-38000 Grenoble, France\\
\email{\{romain.xu-darme,zakaria.chihani\}(at)cea.fr} \\
\email{\{georges.quenot,marie-christine.rousset\}(at)imag.fr}
}

\maketitle
\begin{abstract}
In this paper, we present PARTICUL, a novel algorithm for unsupervised learning of part detectors from datasets used in fine-grained recognition. It exploits the macro-similarities of all images in the training set in order to mine for recurring patterns in the feature space of a pre-trained convolutional neural network. We propose new objective functions enforcing the locality and unicity of the detected parts. Additionally, we embed our detectors with a confidence measure based on correlation scores, allowing the system to estimate the visibility of each part.
We apply our method on two public fine-grained datasets (Caltech-UCSD Bird 200 and Stanford Cars) and show that our detectors can consistently highlight parts of the object while providing a good measure of the confidence in their prediction. We also demonstrate that these detectors can be directly used to build part-based fine-grained classifiers that provide a good compromise between the transparency of prototype-based approaches and the performance of non-interpretable methods.
\end{abstract}

\section{Introduction}\label{sec:particul_intro}

With the development of deep learning in recent years, convolutional neural networks (CNNs) have quickly become the backbone of all state-of-the-art visual recognition systems.
CNNs are highly efficient but also highly complex systems manipulating abstract (and often opaque) representations of the image - also called feature vectors - to achieve high accuracy, at the cost of transparency and interpretability of the decision-making process.
In order to overcome these issues, a solution, explored in~\cite{farhadi_describing_2009,lampert_attribute_2014,abdulnabi_multi-task_2015,hendricks_generating_2016,liang_unified_2015,wang_attribute_2017,zhu_multi-label_2017,fukui_attention_2019,han_attribute-aware_2019,chen_this_2019,hassan_explaining_2019,zhao_recognizing_2019,nauta_neural_2021,rymarczyk_protopshare_2020},
consists in building an intermediate representation associating each input image with a set of semantic \textit{attributes}. These attributes, usually representing an association between a part of an object (\eg head of a bird) and a property (\eg shape, color), can be either used by interpretable methods~\cite{chen_this_2019,nauta_neural_2021,rymarczyk_protopshare_2020} to produce the decision, as \textit{post-hoc} explanation~\cite{hendricks_generating_2016,hassan_explaining_2019} of a particular decision, or as supplementary information in order to discriminate similar categories in the case of fine-grained visual classification (FGVC)~\cite{zhu_multi-label_2017,fukui_attention_2019,han_attribute-aware_2019,zhao_recognizing_2019,liang_unified_2015}.
In practice, attributes are learned through fully supervised algorithms~\cite{farhadi_describing_2009,wang_attribute_2017,zhu_multi-label_2017,fukui_attention_2019,han_attribute-aware_2019,zhao_recognizing_2019,liang_unified_2015,abdulnabi_multi-task_2015}, using datasets with hand-crafted annotations. Such datasets are expensive to produce - using expert knowledge or online crowd-sourcing platforms - and prone to errors~\cite{jo_lessons_2020,northcutt_pervasive_2021}. Therefore, a lot of effort~\cite{korsch_classification-specific_2019,chen_this_2019,ge_weakly_2019,peng_object_part_2018,yang_learning_2018,ding_selective_2019,li_attribute_2020,zhou_human-understandable_2021,zheng_learning_2017,sicre_unsupervised_2017,zhang_unsupervised_2019,han_pcnn_2022} has been recently put into the development of more scalable techniques using less training information. In particular, the task of localizing different parts of an object, a prerequisite to attribute detection, can be performed in a weakly supervised (using the category of each image) or unsupervised manner.
By focusing on general features of the objects rather than discriminative details, part detection represents an easier task than attribute detection. However, it must offer strong evidence of accuracy and reliability in order to constitute a solid basis for a trustworthy decision.

In this paper, we present PARTICUL\footnote{Patent pending} (Part Identification with Confidence measure using Unsupervised Learning), a plug-in module that uses an unsupervised algorithm in order to learn part detectors from FGVC datasets. It exploits the macro-similarities of all images in the training set in order to mine for recurring patterns in the feature space of a pre-trained CNN. We propose new objective functions enforcing the locality and unicity of the detected parts. Our detectors also provide a confidence measure based on correlation scores, allowing the system to estimate the visibility of each part. We apply our method on two public datasets, Caltech-UCSD Bird 200 (CUB-200)~\cite{welinder_CUB200_2010} and Stanford Cars~\cite{yang_large_scale_2015}, and show that our detectors can consistently highlight parts of the object while providing a good measure of the confidence in their prediction. Additionally, we provide classification results in order to showcase that classifiers based only on part detection can constitute a compromise between accuracy and transparency.

This paper is organized as follows: Section~\ref{sec:particul_sota} presents the related work on part detection; Section~\ref{sec:particul_alg} describes our PARTICUL model and Section~\ref{sec:particul_exps} presents our results on two FGVC datasets. Finally, Section~\ref{sec:particul_conclusion} concludes this paper and proposes several lines of research aiming at improving our approach.

\section{Related work}\label{sec:particul_sota}
Part detection (and more generally attribute detection) is a problem which has been extensively studied in recent years, especially for FGVC which is a notoriously hard computer vision task.
Learning how to detect object parts in this context
can be done either in a fully supervised
 (using ground-truth part locations), weakly supervised (using image labels only) or unsupervised manner.

Weakly supervised approaches~\cite{kun_duan_discovering_2012,xiao_application_2014,simon_neural_2015,liu_fine-grained_2017,korsch_classification-specific_2019,chen_this_2019,ge_weakly_2019,peng_object_part_2018,yang_learning_2018,ding_selective_2019,li_attribute_2020,zhou_human-understandable_2021,zheng_learning_2017} produce part detectors as by-products of image classification \ie object parts are jointly learned with categories in an end-to-end manner to help distinguish very similar categories. %
In particular, the OPAM approach presented in \cite{peng_object_part_2018} obtains parts from candidate image patches that are generated by Selective Search~\cite{Uijlings2013SelectiveSF}, filtered through a dedicated pre-trained network, selected using an object-part spatial constraint model taking into account a coarse segmentation of the object at the image level, and finally semantically realigned by applying spectral clustering on their corresponding convolutional features.

Recently, unsupervised part detection methods have greatly benefited from the expressiveness and robustness of features extracted from images using deep CNNs. Modern approaches~\cite{krause_fine-grained_2015,sicre_unsupervised_2017,zhang_unsupervised_2019,zheng_learning_2017,zhang_picking_2016} mainly focus on applying clustering techniques in the feature space and/or identifying convolutional channels with consistent behaviors.
More precisely, \cite{krause_fine-grained_2015} produces part detectors through the sampling and clustering of keypoints within an estimation of the object segmentation produced by a method similar to GrabCut~\cite{rother_grabcut_2004}. For a given image, \cite{zhang_unsupervised_2019} uses itemset mining to regroup convolutional channels based on their activation patterns, producing parts per image instead of globally (as in \cite{peng_object_part_2018}) and thus requiring an additional semantic realignment - through clustering - for part-based classification.
Rather than working on the full images, \cite{zhang_picking_2016} and \cite{sicre_unsupervised_2017} use convolutional features of region proposals produced by Selective search~\cite{van_de_sande_segmentation_2011}.
\cite{sicre_unsupervised_2017} then learns the relation between regions and parts using clustering and soft assignment algorithms, while~\cite{zhang_picking_2016} builds part detectors from the top-$k$ most activated filters across all regions of the training set.
Due to the independent training of each detector, the latter approach produces tens of redundant part detectors which must be filtered out in a weakly unsupervised manner. %

This issue is partially addressed in the MA-CNN approach~\cite{zheng_learning_2017}, where part detectors are learned by performing a soft assignment of the convolutional channels of a pretrained CNN into groups that consistently have a peak response in the same neighborhood, then regressing the weight (importance) of each channel in each group through a fully convolutional layer applied on the feature map. Each part detector produces an activation map, normalized using a sigmoid function, corresponding to the probability of presence of the part at each location. By picking the location with the highest activation value for each detector, MA-CNN generates part-level images patches that are fed into a Part-CNN~\cite{branson_bird_2014} architecture for classification.
In practice, part detectors are initialized by channel grouping (using k-means clustering) and pre-trained in an unsupervised manner using dedicated loss functions enforcing the locality and unicity of each detector attention region. Finally, these detectors are fine-tuned during end-to-end learning of the image category. As such, this method falls into the category of unsupervised part learning with weakly supervised fine-tuning.
More recently, the P-CNN approach~\cite{han_pcnn_2022} implements the part detection learning algorithm of MA-CNN, replacing the fully connected layer in charge of channel grouping by a convolution layer. Its classification pipeline uses Region-of-Interest (ROI) pooling layer around the location of maximum activation to directly extract part features and global features from the part detection backbone.
In both cases, the channel grouping initialization requires first to process all images of the training set in order to cluster channels according to the location of their highest response.
Moreover, for a given detector, only the area immediately surrounding the location of highest activation is taken into account (either to extract an image patch in~\cite{zheng_learning_2017} or a vector through ROI-pooling in~\cite{han_pcnn_2022}), thus reducing the ability to detect the same part at different scales.

\paragraph{Contributions} Our PARTICUL approach builds part detectors as weighted sums of convolutional channels extracted from a pre-trained CNN backbone. Our detectors are trained globally across the training set, instead of locally, contrary to~\cite{zhang_unsupervised_2019} and \cite{peng_object_part_2018} which require an additional semantic alignment phase.
We develop objective functions that are specially crafted to ensure the compactness of the distribution of activation values for each part detector and the diversity of attention locations. Unlike~\cite{zheng_learning_2017} and~\cite{han_pcnn_2022}, which propose similar functions, our approach does not require any fine-tuning of the backbone or an initial channel grouping phase. This induces a fast convergence during training and enables our module to be used with black-box backbones, in a plug-in fashion. Moreover, our detectors use a softmax normalization function, instead of sigmoid function, that simplifies the process of locating the part and enables us to detect parts at different scales.
Finally, our detectors supply a measure of confidence in their decision based on the distribution of correlation scores across the training set. Importantly, this measure can also predict the \emph{visibility} of a given part, a subject which is, to our knowledge, not tackled in any of the related work.

\section{Proposed model}\label{sec:particul_alg}
In this section, we present PARTICUL, our proposal for unsupervised part learning from FGVC datasets.
In practice, each part detector can be seen as a function highlighting dedicated attention regions on the input image, in accordance with
the constraints detailed in this section.

\paragraph{Notations}
For a tensor $T$ inside a model $\mathcal{M}$, we denote $T(x)$ the value of $T$ given an input $x$ of $\mathcal{M}$. If $T$ has $H$ rows, $W$ columns and $C$ channels, we denote $\featvec{T}{h}{w}(x)\in \mathbb{R}^C$ the vector (or single feature) located at the $h^{th}$ row, $w^{th}$ column.
We denote $*$ the convolution operation between tensors, $\sigma$ the softmax normalization function.
e define $\mathcal{I}$ as the set of all input images of a given dimension $H_\mathcal{I}\times W_\mathcal{I}$ and $\mathcal{I}_\mathcal{C}\subseteq \mathcal{I}$ as the subset of images in $\mathcal{I}$ containing an object of the macro-category $\mathcal{C}$. In our experiments, we denote $X_{train}=\big\{x_i\in \mathcal{I}_\mathcal{C} \big| i\in [1\isep n]\big\}$ the training set of size $n$. Although $\mathcal{I}_\mathcal{C}$ cannot usually be formally specified, we assume that $X_{train}$ is representative of this set.

\subsection{Part detectors}
\begin{figure}[htbp]
\centering
    \includegraphics[width=0.90\textwidth]{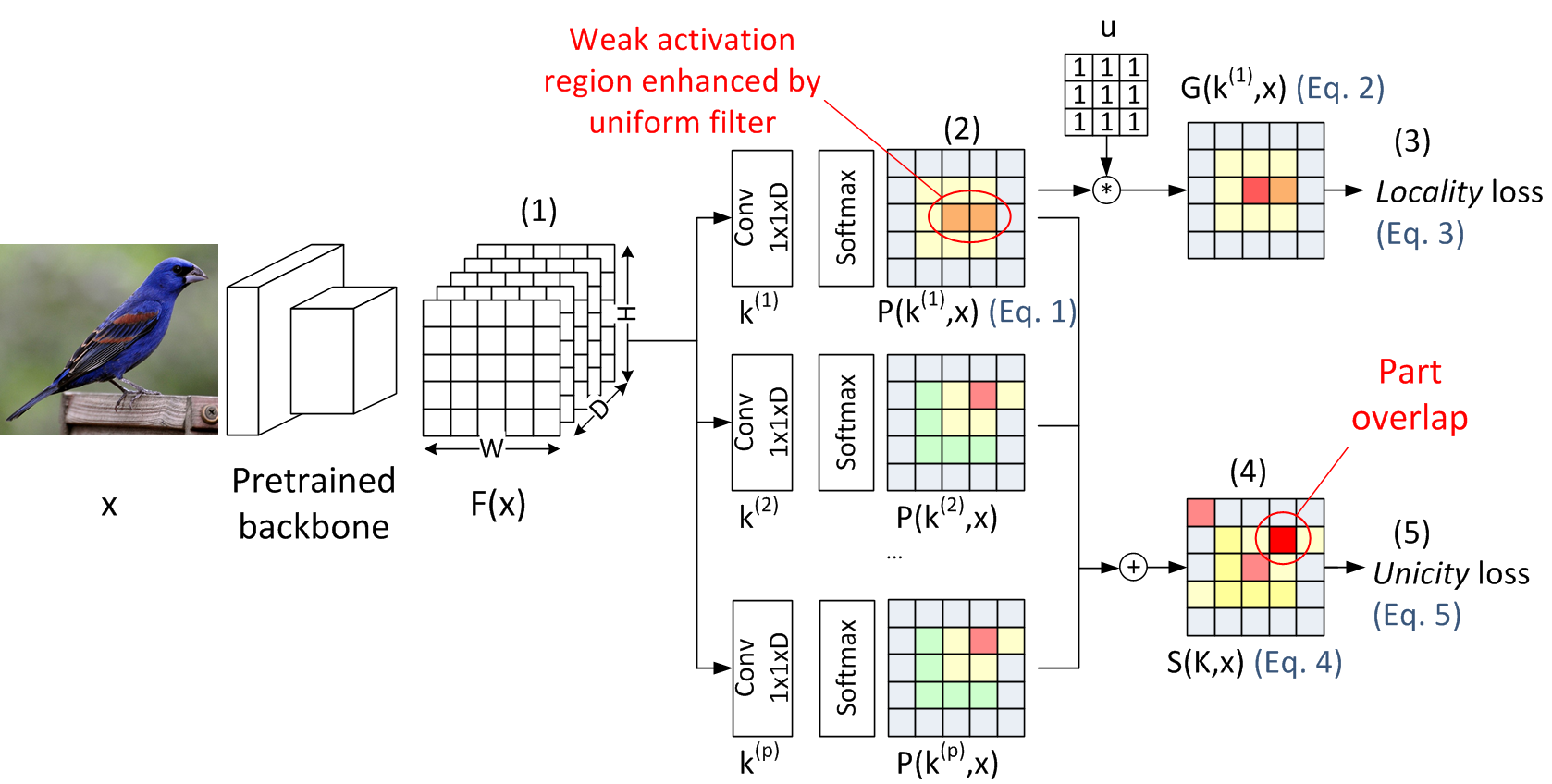}
    \caption{Architecture of our PARTICUL model. (1)~Convolutional features $F(x)$ are extracted from the image $x$. (2)~Each part detector produces an activation map $\partmap{i}{x}$. (3)~We apply a uniform kernel $u$ to each part activation map before computing the Locality loss which ensures the compactness of activations. (4) All part activation maps are summed in $S(K,x)$. (5) Unicity loss is applied to ensure the diversity of part detectors. Here, part detectors $2$ and $p$ are very similar, leading to a high peak in bright red in $S(K,x)$. Best viewed in color.}
    \label{fig:particul_arch}
\end{figure}

As illustrated in Fig.~\ref{fig:particul_arch}, we first use a pre-trained CNN $F$ to extract a $H\times W \times D$ feature map. Then, we build each of the $p$ part detectors as a convolution with a $1\times1\times D$ kernel $k^{(i)}$ (no bias is added), followed by a 2D spatial softmax layer, each part detector producing a $H\times W$ activation map
\begin{equation}\label{eq:Pi}
    \partmap{i}{x} = \sigma\big(F(x)*k^{(i)}\big),~\forall x\in \mathcal{I}
\end{equation}
We denote the set of all part detector kernels as $K=[k^{(1)},k^{(2)},\ldots, k^{(p)}]$.

\subsection{Objective functions}
\paragraph{Locality:}
For each part detector $i$, we force the network to learn a convolutional kernel $k^{(i)}$ which maximizes one region of the activation map for any training image $x$. In order to allow activations to be localized into a given neighborhood rather than in a single location in the $H\times W$ activation map, we first apply a $3\times 3$ uniform filter $u$ on $\partmap{i}{x}$. Let
\begin{equation}\label{eq:Gi}
  G(k^{(i)},x)=\partmap{i}{x}*\underbrace{
    \begin{pmatrix}
        1 & 1 & 1 \\
        1 & 1 & 1 \\
        1 & 1 & 1
    \end{pmatrix}}_{u}
\end{equation}

Learning all convolutional kernels $K$ in order to enforce locality can be translated as the optimization of the objective function
\begin{equation}\label{eq:loss_locality}
\mathcal{L}_{l}(K)=-\dfrac{1}{p} \sum\limits_{i=1}^p\dfrac{1}{n}\sum \limits_{x\in X_{train}}  \max\limits_{h,w}\Big(\featvec{G(k^{(i)},x)}{h}{w}\Big)
\end{equation}
Intuitively, solving the optimization problem in Eq.~\ref{eq:loss_locality} using $G(k^{(i)},x)$ rather than $\partmap{i}{x}$ relaxes the learning constraint and prevents the detectors from focusing on discriminative details between two adjacent feature vectors that would represent the same part.

\paragraph{Unicity} In order to prevent the system from learning only a handful of easy parts, we wish to ensure that each feature vector $\featvec{F}{h}{w}(x)$ is not simultaneously correlated with multiple convolutional kernels $k^{(i)}$. Let $S(K,x)\in \mathbb{R}^{H\times W}$ s.t.:
\begin{equation}\label{eq:particul_sum}
 \featvec{S}{h}{w}(K,x)=\sum\limits_{i=1}^p \featvec{\partmap{i}{x}}{h}{w}
\end{equation}
Ensuring unicity can be translated as the following objective function, \ie making sure that no location $(h,w)$ contains a cumulative activation higher than $1$:
\begin{equation}\label{eq:loss_unicity}
    \mathcal{L}_{u}(K)=\dfrac{1}{n}\sum\limits_{x\in X_{train}}\max\Big(\max\limits_{h,w}\featvec{S}{h}{w}(K,x)-1,0\Big)
\end{equation}

The final objective function for the unsupervised learning of part detectors is a weighted composition of the functions described above:
\begin{equation}\label{eq:part_detect}
    \mathcal{L}(K) = \mathcal{L}_{l}(K)+\lambda \mathcal{L}_{u}(K)
\end{equation}
where $\lambda$ controls the relative importance of each objective function.

\subsection{Confidence measure and visibility}\label{sec:filter_correlation}
Visibility is related to a measure of confidence in the decision provided by each of our detectors and based on the distribution of correlation scores across the training set. After fitting our detectors to Eq.~\ref{eq:part_detect}, we employ the function
\begin{equation}\label{eq:Hi}
\begin{array}{lll}
    H_i:& \mathcal{I} \rightarrow & \mathbb{R} \\
     & x \rightarrow & \max\limits_{h,w} \big( \featvec{F}{h}{w}(x)*k^{(i)}\big)
\end{array}
\end{equation}
returning the maximum correlation score of detector $i$ for image $x$ (before softmax normalization). The distribution of values taken by $H_i$ across $\mathcal{I}_\mathcal{C}$ is modeled as a random variable following a normal distribution $\mathcal{N}(\mu_i,\sigma_i^2)$ estimated over $X_{train}$.
We define the confidence measure of part detector $i$ on image $x$ as
\begin{equation}\label{eq:confidence_measure}
    C(x,i) =
    \Phi(H_i(x),\mu_i,\sigma_i^2)
\end{equation}
where $\Phi(z,\mu,\sigma^2)$ is the cumulative distribution function of $\mathcal{N}(\mu_i,\sigma_i^2)$.

\section{Experiments}\label{sec:particul_exps}
In order to showcase the effectiveness of our approach, we apply our algorithm on two public FGVC datasets - the Caltech-UCSD Birds 200 (CUB-200)~\cite{welinder_CUB200_2010} dataset containing 11,788 images from 200 bird species (5994 training images, 5794 test images) and the Stanford Cars~\cite{yang_large_scale_2015} dataset containing 16,185 images from 196 car models (8144 training images, 8041 test images).
In addition to the object subcategory labels (bird species, car model), both datasets also provide additional information in the form of annotations (object bounding box, part locations in \cite{welinder_CUB200_2010}). However, when learning and calibrating our part detectors, and unless specified otherwise, \textbf{we do not use any information other than the images themselves} and work in a fully unsupervised setting.

\subsection{Unsupervised learning of part detectors}
In this section, we illustrate our unsupervised part detection algorithm using VGG19~\cite{Simonyan2015Very} (with batch normalization) pretrained on the Imagenet dataset~\cite{deng_imagenet_2009} as our extractor $F(.)$ (Eq.~\ref{eq:Pi}). For part visualization, we use SmoothGrad~\cite{Smilkov2017SmoothGradRN}, filtering out small gradients using the method described in~\cite{Otsu79Threshold}.

 We train our detectors during 30 epochs, using RMSprop with a learning rate of $5\times 10^{-4}$ and a decay of $10^{-5}$. These training parameters are chosen using cross-validation on the training set, with the goal of minimizing the objective function (Eq.~\ref{eq:part_detect}). Since we do not need to fine-tune the extractor, only $p\times D$ convolutional weights are learned during training, which drastically reduces the computation time. Finally, for a given number of parts $p$, we supervise the importance $\lambda$ of the unicity constraint (Eq.~\ref{eq:part_detect}) by measuring the overall attention across the feature map after training. More precisely, we compute the value
\begin{equation}\label{eq:sum_max_attention}
	\mathcal{E}(X_{train},K) = \dfrac{1}{|X_{train}|}\sum\limits_{x\in X_{train}} \dfrac{1}{p} \sum\limits_{h,w} \max\limits_{i\in [1 \isep p]} \featvec{\partmap{i}{x}}{h}{w}
\end{equation}
corresponding to the average contribution of each detector. $\mathcal{E}(X_{train},K) = 1$ corresponds to the ideal case where all detectors focus on different locations in the feature map $F(x)$, while $\mathcal{E}(X_{train},K) = 1/p$ indicates a high redundancy of attention regions among detectors.
\begin{figure}[htbp]
	\centering
  \begin{subfigure}[b]{0.48\textwidth}
		\centering
		\includegraphics[width=\textwidth]{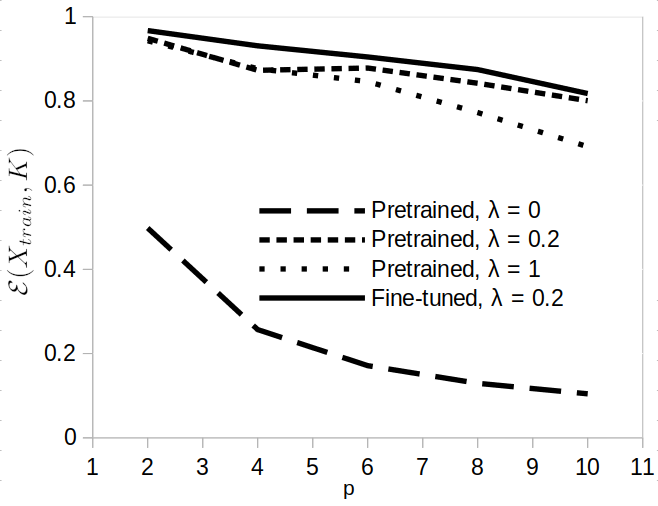}
		\caption{CUB-200}
    \label{fig:particul_cub200_energy}
	\end{subfigure}
	\hfill
  \begin{subfigure}[b]{0.48\textwidth}
	  \centering
		\includegraphics[width=\textwidth]{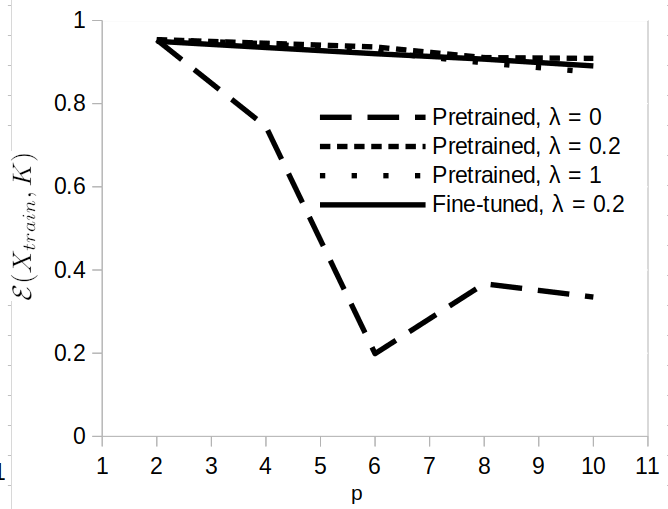}
    \caption{Stanford Cars}
    \label{fig:particul_cars_energy}
  \end{subfigure}
  \hfill
	\caption{$\mathcal{E}(X_{train},K)$ v. number $p$ of detectors for various values of $\lambda$, with pre-trained or fine-tuned extractor.}
	\label{fig:energy}
\end{figure}
As illustrated in Fig.~\ref{fig:energy}, for both datasets the average contribution of each detector decreases with $p$, reflecting the growing difficulty of finding distinct detectors. The choice of $p$ itself depends on the downstream task (\eg classification), where again increasing the number of parts might produce diminishing returns (see Sec.~\ref{sec:classification}). Moreover, in both datasets the presence ($\lambda > 0$) of our unicity loss function (Eq.~\ref{eq:loss_unicity}) is paramount to learning distinct detectors. Without the unicity constraint, in the case of CUB-200 all detectors systematically converge towards the same location inside of the feature map ($\mathcal{E}(X_{train},K)\approx 1/p$) corresponding to the head of the bird; in the case of Stanford Cars, where images contain more varied distinctive patterns, the number of unique detectors depends on the random initialization of their convolutional weights, but the average contribution per detector quickly drops with $p$.
For both datasets, we choose $\lambda=0.2$ which maximizes $\mathcal{E}(X_{train},K)$ for all values of $p$. As a comparison, we also train our detectors after fine-tuning the extractor on each dataset (in this case, the learning process can be considered weakly supervised), leading to only marginally better results. This supports our claim that, in practice, our detectors do not require a fine-tuned extractor and can work as a plug-in module to a black-box extractor.

\begin{figure}[htbp]
		\centering
		\includegraphics[width=0.84\textwidth]{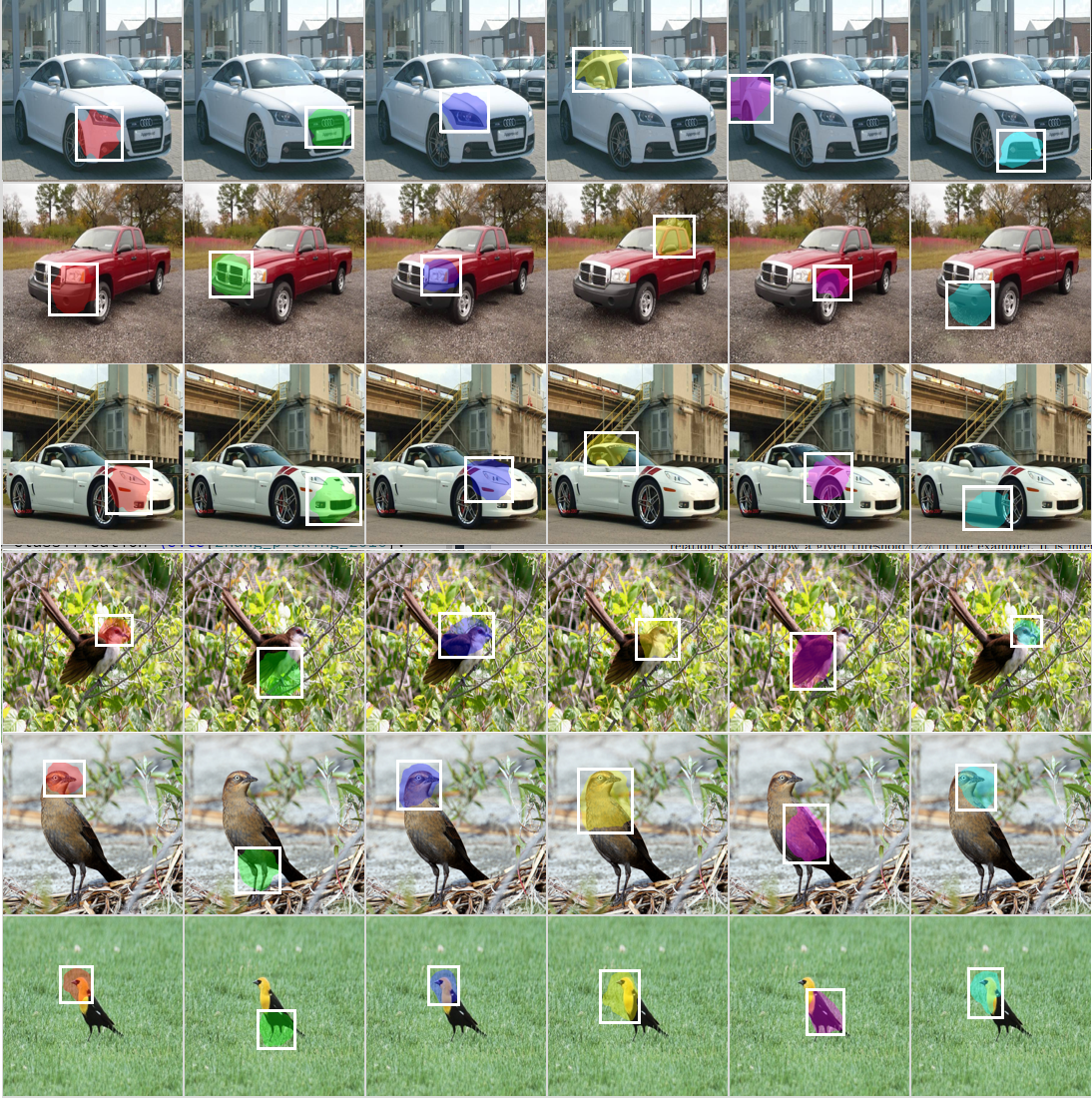}
		\caption{Part visualization after training 6 detectors using the VGG19 extractor on Stanford cars (top) or CUB-200 (bottom). Using this visualizations, we can manually re-attach a semantic value to each detector, \eg the second part detector trained on CUB-200 is probably a "leg" detector. Best viewed in color.}
    \label{fig:particul_parts}
\end{figure}
As illustrated in Fig.~\ref{fig:particul_parts}, our detectors consistently highlight recurring parts of the objects and are relatively insensitive to the scale of the part. Although we notice some apparent redundancy of the detected parts
(\eg around the head of the bird), the relatively high corresponding value for $\mathcal{E}(X_{train},K)$ (0.87 for $\lambda=0.2$ and $p=6$, see Fig.~\ref{fig:particul_cub200_energy}) indicates that each detector actually focus on a different location of the feature map, leading to a richer representation of the object.
Note that when training our detectors, we do not know beforehand towards which part each detector will converge. However, using part visualization, we can re-attach a semantic value to each detector after training.

\subsection{Confidence measure}
After training our detectors, we perform a calibration of their confidence measure using the method presented in Sec.~\ref{sec:filter_correlation}. In order to illustrate the soundness of our approach, we exploit the annotations provided by the CUB-200 dataset to extract a subset of images where the legs of the bird are non-visible (2080 images from both the training and test set, where the annotations indicate that the color of the legs is not visible).
\begin{figure}[htbp]
	\centering
  \begin{subfigure}[b]{0.45\textwidth}
		\centering
		\includegraphics[width=\textwidth]{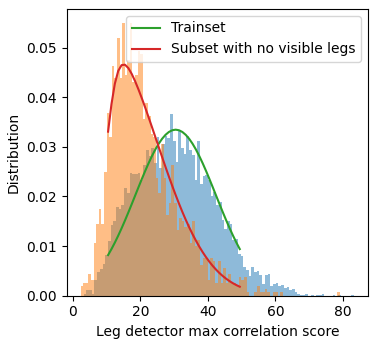}
		\caption{Distribution of maximum correlation scores on the CUB-200 training set (in blue) and on a subset containing only images with non-visible legs (red).}
    \label{fig:particul_legs_distrib}
	\end{subfigure}
	\hfill
  \begin{subfigure}[b]{0.45\textwidth}
	  \centering
    \includegraphics[width=\textwidth]{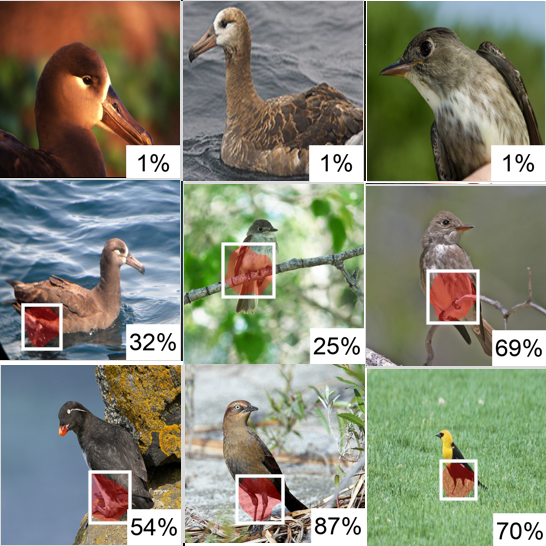}
    \caption{Confidence scores and part visualizations on images with non visible legs (top-row) and with visible legs (bottom rows).}
    \label{fig:particul_legs}
  \end{subfigure}
  \hfill
  \caption{Confidence measure applied on a bird leg detector trained and calibrated on CUB-200 dataset. Best viewed in color.}
\end{figure}
As shown on Fig.~\ref{fig:particul_legs_distrib}, there is a clear difference in the distributions of maximum correlation scores between images with and without visible legs. This also confirms that images not containing the part tend to produce lower correlation scores, and by consequence our proposal for a confidence measure based on a cumulative distribution function (Eq. \ref{eq:confidence_measure}).
In practice (Fig.~\ref{fig:particul_legs}), a calibrated detector can detect the same part at different scales while ignoring images where the confidence measure is below a given threshold (2\% in the example). It is interesting to note that for all images in the middle row, the annotations actually indicate that the legs are not visible, \ie our detectors can also be employed as a fast tool to verify manual annotations.

\subsection{Classification}\label{sec:classification}
In all related works closest to ours, the performance of part detectors is never evaluated independently but rather \wrt to the classification task, a method which is highly dependent on the architecture chosen for the decision-making process: \eg SVMs in \cite{krause_fine-grained_2015} and \cite{sicre_unsupervised_2017}, Part-CNN~\cite{branson_bird_2014} in \cite{zheng_learning_2017}, Spatially Weighted Fisher Vector CNN (SWFV) in \cite{zhang_picking_2016}, multi-stream architecture in \cite{zhang_unsupervised_2019}. Nevertheless, in order to provide a basis of comparison with state-of-art techniques, we also provide classification results based on the extraction of part feature vectors.
\begin{figure}[htbp]
\centering
    \includegraphics[width=\textwidth]{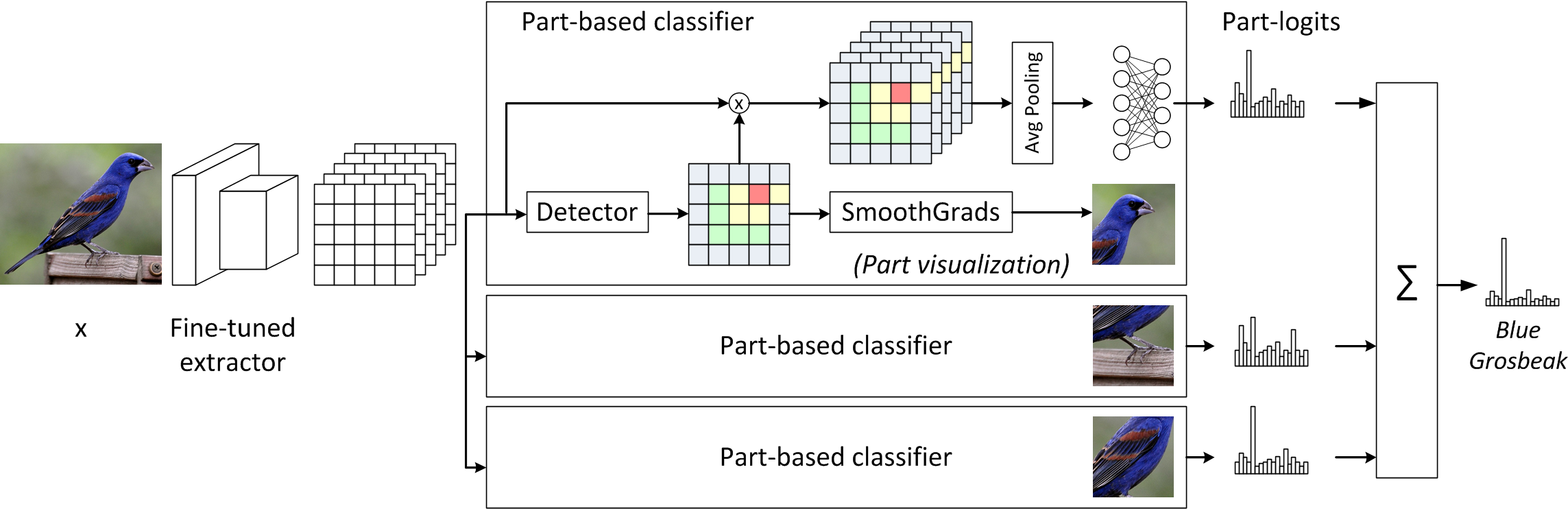}
    \caption{Our classification model for FGVC datasets. The feature map produced by the extractor is masked out using the activation map of each detector and processed through a pooling layer, resulting in a set of part vector features. These vectors are processed independently through a set of fully connected layers to generate part-based logits that are summed up to produce the final prediction.}
    \label{fig:particul_cnet}
\end{figure}
As illustrated in Fig.~\ref{fig:particul_cnet}, we use a method similar to \cite{zhao_recognizing_2019}, where the feature map $F(X)$ is multiplied element-wise with the activation map produced by each detector to compute a set of part feature vectors that are used as inputs of $p$ independent classifiers (each containing 2 intermediate layers - with 4096 neurons, ReLU activation and dropout - before the classification layer). As such, our detectors operate a form of semantic realignment and extraction of relevant feature vectors from the image. Note that, contrary to the fixed size ROI pooling used in \cite{han_pcnn_2022}, this method has the advantage of dynamically adjusting the number of feature vectors taken into account depending on the scale of the part. For the final decision, we sum up the logits of all part-based classifiers to produce prediction scores for each category. Therefore, the final decision can be directly traced back to the individual result of each part-based classifier.
This approach does not provide the transparency~\cite{Lipton2018TheMO} of prototype-based methods~\cite{chen_this_2019,nauta_neural_2021}, but constitutes a good compromise to non interpretable approaches using global features.
\begin{table*}[]
	\setlength{\tabcolsep}{5pt}
	\renewcommand{\arraystretch}{1}
	\newcommand{\STAB}[1]{\begin{tabular}{@{}c@{}}#1\end{tabular}}
	\newcommand*{\MyIndent}{\hspace*{0.5cm}}%
	\footnotesize
	\centering
	\caption{Classification accuracy on CUB-200 and Stanford Cars and comparison with related works, from less transparent to most transparent.}
	\begin{tabular}{cl|c|l|l}
		& \multirow{2}{*}{Method} & Train & \multicolumn{2}{c}{Accuracy (\%)} \\
		\cline{4-5}
		& & anno. &  CUB-200 & Stanford Cars \\
		\cline{2-5}
		\multirow{20}{*}{\STAB{\rotatebox[origin=cl]{90}{\bigskip $\longleftarrow$ Transparency\cite{Lipton2018TheMO} $\longleftarrow$}}}
		& \textbf{Global features only} & & & \\
		& \MyIndent Baseline (VGG-19) & & 83.3 & 89.3 \\
		\cline{2-5}
		& \textbf{Part + global features} & & & \\
		& \MyIndent UPM\cite{zhang_unsupervised_2019} (VGG19) &  & 81.9 & 89.2 \\
		& \MyIndent OPAM~\cite{peng_object_part_2018} (VGG-16) & & 85.8 & 92.2 \\
		& \MyIndent MA-CNN~\cite{zheng_learning_2017} (VGG-19) & & 86.5 & 92.8 \\
		& \MyIndent P-CNN~\cite{han_pcnn_2022} (VGG-19) & & 87.3 & 93.3 \\
		\cline{2-5}
		& \textbf{Part-based} & & & \\
		& \MyIndent Ours (VGG-19) & &  &  \\
		& \MyIndent \MyIndent 2 parts & & 70.1  & 76.0 \\
		& \MyIndent \MyIndent 4 parts & & 79.2 & 84.2 \\
		& \MyIndent \MyIndent 6 parts & & 81.5 & 87.5 \\
		& \MyIndent \MyIndent 8 parts & &  82.3 & 88.3 \\
		& \MyIndent \MyIndent 10 parts & &  82.3 & 88.6 \\
		\cline{3-5}
		& \MyIndent OPAM~\cite{peng_object_part_2018} (parts-only, VGG-16) & & 80.7 & 84.3 \\
		& \MyIndent No parts\cite{krause_fine-grained_2015} (VGG19) & BBox & 82.0 & 92.6 \\
		& \MyIndent \MyIndent + Test Bbox    & BBox & 82.8 & 92.8 \\
		& \MyIndent PDFS\cite{zhang_picking_2016} (VGG-19) &  & 84.5 & n/a \\
		\cline{2-5}
		& \textbf{Prototypes} & & & \\
		& \MyIndent ProtoPNet~\cite{chen_this_2019} (VGG-19) & BBox & 78.0 $\pm$ 0.2 & 85.9 $\pm$ 0.2 \\
		& \MyIndent ProtoTree~\cite{nauta_neural_2021} (Resnet-50) &  & 82.2 $\pm$ 0.7 & 86.6 $\pm$ 0.2 \\
	\end{tabular}
	\label{tab:accuracy}
\end{table*}

\begin{figure}[htbp]
\centering
    \includegraphics[width=0.95\textwidth]{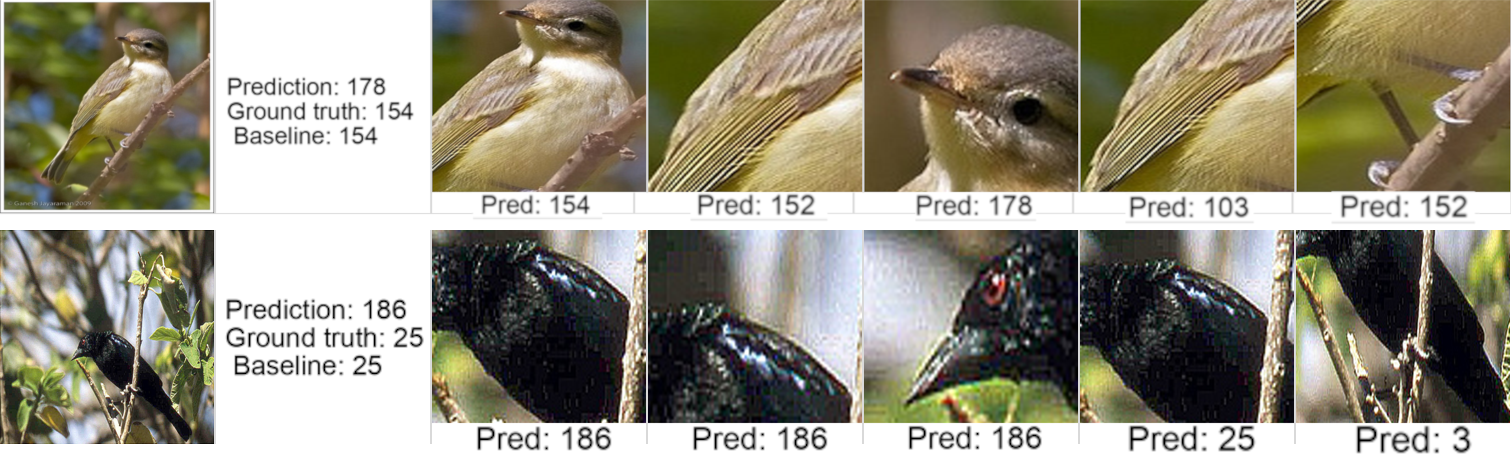}
    \caption{Examples of images correctly classified by the baseline and incorrectly classified by our model (only 5 parts are shown for clarity). In both cases, at least one part-based classifier provides a correct prediction. Best viewed in color.}
    \label{fig:particul_cnet_fails}
\end{figure}

Table ~\ref{tab:accuracy} summarizes the results obtained when using our part-based classifier, along with the results obtained on a fine-tuned VGG-19 (acting as a baseline) and other state-of-the-art methods. With 8 detectors, we outperform the prototype-based approach~\cite{nauta_neural_2021} - which uses a more efficient extractor (Resnet50) - on Stanford Cars and obtain similar results on CUB-200. When compared with other methods using only part-level features, our PARTICUL model outperforms the OPAM~\cite{peng_object_part_2018} approach on both datasets. We also obtain comparable results to other methods requiring either the object bounding box~\cite{krause_fine-grained_2015} or to pre-select detectors (in a weakly supervised manner) based on their classification accuracy~\cite{zhang_picking_2016}. When compared with less transparent methods using image-level (global) features in addition to part-level features - a method which usually has a significant impact on accuracy (\eg from 80.7\% to 85.8\% on CUB-200 using OPAM~\cite{peng_object_part_2018}) - again we achieve comparable results and even outperform the UPM~\cite{zhang_unsupervised_2019} approach on CUB-200.
Finally, it is also interesting to note that for both datasets, our classification results obtained by picking 10 feature vectors out of the $14\times 14=196$ possible vectors of the extractor feature map correspond to a drop of less then 1\% in accuracy when compared with the baseline, indicating that we are indeed selecting the most relevant vectors for the classification.
Moreover, in 80\% (resp. 77\%) of these cases where only the baseline provides a correct prediction on the CUB-200 (resp. Stanford Cars) dataset, at least one of our individual part-based classifier does provide a correct prediction (see Fig.~\ref{fig:particul_cnet_fails}). Thus, our proposed model could be further improved by fine-tuning the relative importance of part logits (\eg using the confidence measure).

\section{Conclusion and future work}\label{sec:particul_conclusion}
In this paper, we presented our algorithm for unsupervised part learning using datasets for FGVC. We showed that our detectors can consistently highlight parts of an object while providing a confidence measure associated with the detection. To our knowledge, our method is the first to take the visibility of parts into account, paving the road for a solid attribute learning and ultimately for interpretable visual recognition. In the particular context of FGVC, our detectors can be integrated in a part-based classification architecture which constitutes a good compromise between the transparency of prototype-based approaches and the performance of non-interpretable methods. As a future work, we will study the integration of our detectors into a prototype-based architecture, learning prototypes from part feature vectors rather than from the entire image feature map.
We will also study the impact of weighting part logits by the confidence score associated with the detected part on the overall accuracy of the system.

\subsubsection*{\footnotesize Acknowledgements}
\footnotesize{Experiments presented in this paper were carried out using the \href{https://www.grid5000.fr}{Grid'5000} testbed, supported by a scientific interest group hosted by Inria and including CNRS, RENATER and several Universities as well as other organizations. This work has been partially supported by MIAI@Grenoble Alpes, (ANR-19-P3IA-0003) and TAILOR, a project funded by EU Horizon 2020 research and innovation programme under GA No 952215. This work was financially supported by European commission through the SAFAIR subproject of the project SPARTA which has received funding from the European Union’s Horizon 2020 research and innovation programme under GA No 830892, as well as through the CPS4EU project that has received funding from the ECSEL Joint Undertaking (JU) under GA No 826276.}

\bibliographystyle{splncs04}
\bibliography{paper}

\begin{thebibliography}{10}
\providecommand{\url}[1]{\texttt{#1}}
\providecommand{\urlprefix}{URL }
\providecommand{\doi}[1]{https://doi.org/#1}

\bibitem{abdulnabi_multi-task_2015}
Abdulnabi, A.H., Wang, G., Lu, J., Jia, K.: Multi-task {CNN} {Model} for
  {Attribute} {Prediction}. IEEE Transactions on Multimedia  \textbf{17}(11),
  1949--1959 (Nov 2015). \doi{10.1109/TMM.2015.2477680},
  \url{http://arxiv.org/abs/1601.00400}, arXiv: 1601.00400

\bibitem{branson_bird_2014}
Branson, S., Van~Horn, G., Belongie, S., Perona, P.: Bird {Species}
  {Categorization} {Using} {Pose} {Normalized} {Deep} {Convolutional} {Nets}.
  BMVC 2014 - Proceedings of the British Machine Vision Conference 2014  (Jun
  2014), \url{http://arxiv.org/abs/1406.2952}, arXiv: 1406.2952

\bibitem{chen_this_2019}
Chen, C., Li, O., Tao, C., Barnett, A.J., Su, J., Rudin, C.: \textit{This}
  looks like \textit{That}: Deep learning for interpretable image recognition.
  Proceedings of the 33rd International Conference on Neural Information
  Processing Systems p. 8930–8941 (2019)

\bibitem{deng_imagenet_2009}
Deng, J., Dong, W., Socher, R., Li, L.J., Li, K., Fei-Fei, L.: Image{N}et: A
  {L}arge-{S}cale {H}ierarchical {I}mage {D}atabase. In: 2009 IEEE Conference
  on Computer Vision and Pattern Recognition. pp. 248--255. IEEE (2009)

\bibitem{ding_selective_2019}
Ding, Y., Zhou, Y., Zhu, Y., Ye, Q., Jiao, J.: Selective {Sparse} {Sampling}
  for {Fine}-{Grained} {Image} {Recognition}. In: 2019 {IEEE}/{CVF}
  {International} {Conference} on {Computer} {Vision} ({ICCV}). pp. 6598--6607.
  IEEE, Seoul, Korea (South) (Oct 2019). \doi{10.1109/ICCV.2019.00670},
  \url{https://ieeexplore.ieee.org/document/9008286/}

\bibitem{farhadi_describing_2009}
Farhadi, A., Endres, I., Hoiem, D., Forsyth, D.: Describing {O}bjects by their
  {A}ttributes. In: 2009 IEEE Conference on Computer Vision and Pattern
  Recognition. pp. 1778--1785 (2009). \doi{10.1109/CVPR.2009.5206772}

\bibitem{fukui_attention_2019}
Fukui, H., Hirakawa, T., Yamashita, T., Fujiyoshi, H.: Attention {B}ranch
  {N}etwork: Learning of {A}ttention {M}echanism for {V}isual {E}xplanation.
  In: 2019 IEEE/CVF Conference on Computer Vision and Pattern Recognition
  (CVPR). pp. 10697--10706 (2019). \doi{10.1109/CVPR.2019.01096}

\bibitem{ge_weakly_2019}
Ge, W., Lin, X., Yu, Y.: {W}eakly {S}upervised {C}omplementary {P}arts {M}odels
  for {F}ine-{G}rained {I}mage {C}lassification {F}rom the {B}ottom {U}p. 2019
  IEEE/CVF Conference on Computer Vision and Pattern Recognition (CVPR) pp.
  3029--3038 (2019)

\bibitem{han_pcnn_2022}
Han, J., Yao, X., Cheng, G., Feng, X., Xu, D.: P-cnn: Part-based convolutional
  neural networks for fine-grained visual categorization. IEEE Transactions on
  Pattern Analysis and Machine Intelligence  \textbf{44}(2),  579--590 (2022).
  \doi{10.1109/TPAMI.2019.2933510}

\bibitem{han_attribute-aware_2019}
Han, K., Guo, J., Zhang, C., Zhu, M.: Attribute-{A}ware {A}ttention {M}odel for
  {F}ine-grained {R}epresentation {L}earning. Proceedings of the 26th ACM
  international conference on Multimedia  (2018)

\bibitem{hassan_explaining_2019}
Hassan, M.U., Mulhem, P., Pellerin, D., Qu{\'e}not, G.: Explaining {V}isual
  {C}lassification using {A}ttributes. 2019 International Conference on
  Content-Based Multimedia Indexing (CBMI) pp.~1--6 (2019)

\bibitem{hendricks_generating_2016}
Hendricks, L.A., Akata, Z., Rohrbach, M., Donahue, J., Schiele, B., Darrell,
  T.: Generating {V}isual {E}xplanations. In: Computer Vision -- ECCV 2016. pp.
  3--19. Springer International Publishing (2016)

\bibitem{jo_lessons_2020}
Jo, E.S., Gebru, T.: Lessons from {A}rchives: {S}trategies for {C}ollecting
  {S}ociocultural {D}ata in {M}achine {L}earning. In: Proceedings of the 2020
  Conference on Fairness, Accountability, and Transparency. p. 306–316. FAT*
  '20, Association for Computing Machinery (2020).
  \doi{10.1145/3351095.3372829}, \url{https://doi.org/10.1145/3351095.3372829}

\bibitem{korsch_classification-specific_2019}
Korsch, D., Bodesheim, P., Denzler, J.: Classification-{Specific} {Parts} for
  {Improving} {Fine}-{Grained} {Visual} {Categorization}. German Conference on
  Pattern Recognition pp. 62--75 (10 2019)

\bibitem{krause_fine-grained_2015}
Krause, J., Jin, H., Yang, J., Fei-Fei, L.: Fine-{Grained} {Recognition}
  {Without} {Part} {Annotations}. In: 2015 IEEE Conference on Computer Vision
  and Pattern Recognition (CVPR). pp. 5546--5555 (2015).
  \doi{10.1109/CVPR.2015.7299194}

\bibitem{kun_duan_discovering_2012}
{Kun Duan}, Parikh, D., Crandall, D., Grauman, K.: Discovering {L}ocalized
  {A}ttributes for {F}ine-{G}rained {R}ecognition. In: 2012 {IEEE} {Conference}
  on {Computer} {Vision} and {Pattern} {Recognition}. pp. 3474--3481. IEEE,
  Providence, RI (Jun 2012). \doi{10.1109/CVPR.2012.6248089},
  \url{http://ieeexplore.ieee.org/document/6248089/}

\bibitem{lampert_attribute_2014}
Lampert, C.H., Nickisch, H., Harmeling, S.: Attribute-{B}ased {C}lassification
  for {Z}ero-{S}hot {V}isual {O}bject {C}ategorization. IEEE Transactions on
  Pattern Analysis and Machine Intelligence  \textbf{36}(3),  453--465 (2014).
  \doi{10.1109/TPAMI.2013.140}

\bibitem{li_attribute_2020}
Li, H., Zhang, X., Tian, Q., Xiong, H.: Attribute {M}ix: {S}emantic {D}ata
  {A}ugmentation for {F}ine {G}rained {R}ecognition. In: 2020 IEEE
  International Conference on Visual Communications and Image Processing
  (VCIP). pp. 243--246 (2020). \doi{10.1109/VCIP49819.2020.9301763}

\bibitem{liang_unified_2015}
Liang, K., Chang, H., Shan, S., Chen, X.: A {Unified} {Multiplicative}
  {Framework} for {Attribute} {Learning}. In: 2015 {IEEE} {International}
  {Conference} on {Computer} {Vision} ({ICCV}). pp. 2506--2514. IEEE, Santiago,
  Chile (Dec 2015). \doi{10.1109/ICCV.2015.288},
  \url{http://ieeexplore.ieee.org/document/7410645/}

\bibitem{Lipton2018TheMO}
Lipton, Z.C.: The mythos of model interpretability. Communications of the ACM
  \textbf{61},  36 -- 43 (2018)

\bibitem{liu_fine-grained_2017}
Liu, X., Xia, T., Wang, J., Yang, Y., Zhou, F., Lin, Y.: Fine-{Grained}
  {Recognition} with {Automatic} and {Efficient} {Part} {Attention}. [Preprint]
  arXiv:1603.06765 [cs]  (Mar 2017), \url{http://arxiv.org/abs/1603.06765}

\bibitem{nauta_neural_2021}
Nauta, M., van Bree, R., Seifert, C.: Neural {P}rototype {T}rees for
  {I}nterpretable {F}ine-grained {I}mage {R}ecognition. 2021 IEEE/CVF
  Conference on Computer Vision and Pattern Recognition (CVPR) pp. 14928--14938
  (2021)

\bibitem{northcutt_pervasive_2021}
Northcutt, C.G., Athalye, A., Mueller, J.: Pervasive label errors in test sets
  destabilize machine learning benchmarks. ArXiv  \textbf{abs/2103.14749}
  (2021)

\bibitem{Otsu79Threshold}
Otsu, N.: A threshold selection method from gray-level histograms. IEEE
  Transactions on Systems, Man, and Cybernetics  \textbf{9}(1),  62--66 (1979).
  \doi{10.1109/TSMC.1979.4310076}

\bibitem{peng_object_part_2018}
Peng, Y., He, X., Zhao, J.: Object-{Part} {Attention} {Model} for
  {Fine}-grained {Image} {Classification}. IEEE Transactions on Image
  Processing  \textbf{27}(3),  1487--1500 (Mar 2018).
  \doi{10.1109/TIP.2017.2774041}, \url{http://arxiv.org/abs/1704.01740}, arXiv:
  1704.01740

\bibitem{rother_grabcut_2004}
Rother, C., Kolmogorov, V., Blake, A.: “{GrabCut}” — {Interactive}
  {Foreground} {Extraction} using {Iterated} {Graph} {Cuts}. ACM SIGGRAPH 2004
  Papers  (2004)

\bibitem{rymarczyk_protopshare_2020}
Rymarczyk, D., Struski, L., Tabor, J., Zieliński, B.: {ProtoPShare}:
  {Prototype} {Sharing} for {Interpretable} {Image} {Classification} and
  {Similarity} {Discovery}. [preprint] arXiv:2011.14340 [cs]  (Nov 2020),
  \url{http://arxiv.org/abs/2011.14340}

\bibitem{van_de_sande_segmentation_2011}
van~de Sande, K.E.A., Uijlings, J.R.R., Gevers, T., Smeulders, A.W.M.:
  Segmentation as {S}elective {S}earch for {O}bject {R}ecognition. In: 2011
  {International} {Conference} on {Computer} {Vision}. pp. 1879--1886. IEEE,
  Barcelona, Spain (Nov 2011). \doi{10.1109/ICCV.2011.6126456},
  \url{http://ieeexplore.ieee.org/document/6126456/}

\bibitem{sicre_unsupervised_2017}
Sicre, R., Avrithis, Y., Kijak, E., Jurie, F.: Unsupervised {Part} {Learning}
  for {Visual} {Recognition}. In: 2017 {IEEE} {Conference} on {Computer}
  {Vision} and {Pattern} {Recognition} ({CVPR}). pp. 3116--3124. IEEE,
  Honolulu, HI (Jul 2017). \doi{10.1109/CVPR.2017.332},
  \url{http://ieeexplore.ieee.org/document/8099815/}

\bibitem{simon_neural_2015}
Simon, M., Rodner, E.: Neural {A}ctivation {C}onstellations: {U}nsupervised
  {P}art {M}odel {D}iscovery with {C}onvolutional {N}etworks. 2015 IEEE
  International Conference on Computer Vision (ICCV) pp. 1143--1151 (2015)

\bibitem{Simonyan2015Very}
Simonyan, K., Zisserman, A.: Very deep convolutional networks for large-scale
  image recognition. CoRR  \textbf{abs/1409.1556} (2015)

\bibitem{Smilkov2017SmoothGradRN}
Smilkov, D., Thorat, N., Kim, B., Vi{\'e}gas, F.B., Wattenberg, M.: Smoothgrad:
  removing noise by adding noise. ArXiv  \textbf{abs/1706.03825} (2017)

\bibitem{Uijlings2013SelectiveSF}
Uijlings, J.R.R., van~de Sande, K.E.A., Gevers, T., Smeulders, A.W.M.:
  Selective search for object recognition. International Journal of Computer
  Vision  \textbf{104},  154--171 (2013)

\bibitem{wang_attribute_2017}
Wang, J., Zhu, X., Gong, S., Li, W.: Attribute {Recognition} by {Joint}
  {Recurrent} {Learning} of {Context} and {Correlation}. 2017 IEEE
  International Conference on Computer Vision (ICCV) pp. 531--540 (2017)

\bibitem{welinder_CUB200_2010}
Welinder, P., Branson, S., Mita, T., Wah, C., Schroff, F., Belongie, S.,
  Perona, P.: {Caltech-UCSD Birds 200}. Tech. Rep. CNS-TR-2010-001, California
  Institute of Technology (2010)

\bibitem{xiao_application_2014}
Xiao, T., Xu, Y., Yang, K., Zhang, J., Peng, Y., Zhang, Z.: The {Application}
  of {Two}-level {Attention} {Models} in {Deep} {Convolutional} {Neural}
  {Network} for {Fine}-grained {Image} {Classification}. 2015 IEEE Conference
  on Computer Vision and Pattern Recognition (CVPR) pp. 842--850 (2015)

\bibitem{yang_large_scale_2015}
Yang, L., Luo, P., Loy, C.C., Tang, X.: A {Large}-{Scale} {Car} {Dataset} for
  {Fine}-{Grained} {Categorization} and {Veriﬁcation}. 2015 IEEE Conference
  on Computer Vision and Pattern Recognition (CVPR) pp. 3973--3981 (2015)

\bibitem{yang_learning_2018}
Yang, Z., Luo, T., Wang, D., Hu, Z., Gao, J., Wang, L.: Learning to {Navigate}
  for {Fine}-grained {Classification}. In: Computer Vision -- ECCV 2018. pp.
  438--454. Springer International Publishing (2018)

\bibitem{zhang_unsupervised_2019}
Zhang, J., Zhang, R., Huang, Y., Zou, Q.: Unsupervised {Part} {Mining} for
  {Fine}-grained {Image} {Classification}. [preprint] ArXiv
  \textbf{abs/1902.09941} (2019)

\bibitem{zhang_picking_2016}
Zhang, X., Xiong, H., gang Zhou, W., Lin, W., Tian, Q.: Picking {Deep} {Filter}
  {Responses} for {Fine}-{Grained} {Image} {Recognition}. 2016 IEEE Conference
  on Computer Vision and Pattern Recognition (CVPR) pp. 1134--1142 (2016)

\bibitem{zhao_recognizing_2019}
Zhao, X., Yang, Y., Zhou, F., Tan, X., Yuan, Y., Bao, Y., Wu, Y.: Recognizing
  {Part} {Attributes} {With} {Insufficient} {Data}. In: 2019 {IEEE}/{CVF}
  {International} {Conference} on {Computer} {Vision} ({ICCV}). pp. 350--360.
  IEEE, Seoul, Korea (South) (Oct 2019). \doi{10.1109/ICCV.2019.00044},
  \url{https://ieeexplore.ieee.org/document/9009781/}

\bibitem{zheng_learning_2017}
Zheng, H., Fu, J., Mei, T., Luo, J.: Learning {Multi}-attention {Convolutional}
  {Neural} {Network} for {Fine}-{Grained} {Image} {Recognition}. In: 2017
  {IEEE} {International} {Conference} on {Computer} {Vision} ({ICCV}). pp.
  5219--5227. IEEE, Venice (Oct 2017). \doi{10.1109/ICCV.2017.557},
  \url{http://ieeexplore.ieee.org/document/8237819/}

\bibitem{zhou_human-understandable_2021}
Zhou, X., Yin, J., Tsang, I.W.H., Wang, C.: Human-{Understandable} {Decision}
  {Making} for {Visual} {Recognition}. In: PAKDD (2021)

\bibitem{zhu_multi-label_2017}
Zhu, J., Liao, S., Lei, Z., Li, S.: Multi-label {C}onvolutional {N}eural
  {N}etwork {B}ased {P}edestrian {A}ttribute {C}lassification. Image and Vision
  Computing  \textbf{58},  224--229 (2017)

\end{thebibliography}

\end{document}